\documentclass[sigconf,anonymous=false]{acmart}
\AtBeginDocument{%
  }

\usepackage{pifont}
\newcommand{\cmark}{\ding{51}}  
\newcommand{\xmark}{\ding{55}}
\usepackage{makecell}
\usepackage{subcaption}
\usepackage{bm}
\usepackage{amsmath}
\usepackage{graphicx}
\usepackage{multirow}

\setcopyright{acmlicensed}
\copyrightyear{2025}
\acmYear{2025}

\acmBooktitle{AI for Financial Fraud Detection and Prevention }

\begin{document}

\title{FraudTransformer: Time-Aware GPT for Transaction Fraud Detection}

\author{Gholamali Aminian}
\affiliation{%
  \institution{The Alan Turing Institute}
  \city{London}
  \country{UK}
}
\email{gaminian@turing.ac.uk}

\author{Andrew Elliott}
\affiliation{%
  \institution{ Glasgow University}
   \city{Glasgow}
  \country{UK}}
\email{andrew.elliott@glasgow.ac.uk}

\author{Tiger Li}
\affiliation{%
  \institution{HSBC}
  \city{London}
  \country{UK}
}
\email{tiger.li@hsbc.com}

\author{Timothy Cheuk Hin Wong}
\affiliation{%
 \institution{HSBC}
  \city{London}
  \country{UK}}
\email{timothy.cheuk.hin.wong@hsbc.com}

\author{Victor Claude Dehon}
\affiliation{%
  \institution{HSBC}
   \city{London}
  \country{UK}}
\email{victor.claude.dehon@hsbc.com}

\author{{\L}ukasz Szpruch}
\affiliation{%
  \institution{Edinburgh University}
   \city{Edinburgh}
  \country{UK}}
  \email{l.szpruch@ed.ac.uk}

\author{Carsten Maple}
\affiliation{%
  \institution{Warwick University}
  \city{Coventry}
  \country{UK}}
  \email{cm@warwick.ac.uk}

  \author{Christopher Read}
\affiliation{%
  \institution{HSBC}
  \city{London}
  \country{UK}}
  \email{christopherread@hsbc.com}

  \author{Martin Brown}
\affiliation{%
  \institution{HSBC}
  \city{London}
  \country{UK}}
  \email{martin.w.brown@hsbc.com}

\author{Gesine Reinert}
\affiliation{%
  \institution{Oxford University}
  \city{Oxford}
  \country{UK}}
  \email{gesine.reinert@keble.ox.ac.uk}

\author{Mo Mamouei}
\affiliation{%
  \institution{HSBC}
  \city{London}
  \country{UK}}
\email{mo.mamouei@hsbc.com}

\renewcommand{\shortauthors}{Aminian, et al.}

\begin{abstract}
Detecting payment fraud in real‐world banking streams requires models that can exploit both the order of events and the irregular time gaps between them. We introduce FraudTransformer, a sequence model that augments a vanilla GPT-style architecture with (i) a dedicated time encoder that embeds either absolute timestamps or inter-event values, and (ii) a learned positional encoder that preserves relative order. Experiments on a large industrial dataset—tens of millions of transactions and auxiliary events—show that FraudTransformer surpasses four strong classical baselines (Logistic Regression, XGBoost and LightGBM) as well as transformer ablations that omit either the time or positional component. On the held-out test set it delivers the highest AUROC and PRAUC.
\end{abstract}



\received{1 October 2025}

\maketitle

\section{Introduction}
Fraudulent activities, such as financial fraud, e-commerce scams, and identity theft, are becoming increasingly sophisticated, necessitating the development of advanced detection methods~\cite{hilal2022financial}. 

Conventional approaches rely on handcrafted features and while can be effective in some cases, often fail to identify nuanced patterns of fraudulent behaviour, especially in dynamic and high-dimensional datasets. 

There is promise in using an automated feature creation methods such as those provided by a transformer architecture.
Such an architecture should be able to identify novel subtle patterns before even human monitors detect them. However transformer architectures cannot easily detect temporal information \cite{zeng2023transformers},  potentially limiting their utility for 
fraud detection, as this is a time-sensitive application.

In this paper we assess the combination of a GPT architecture with time-encoders to provide time embeddings on a dataset of financial transactions. 
 The time-encoders used here are rotational and sinusoidal embeddings combined with a positional encoder; the embeddings can be based on absolute time, or on the relative time between events. We
evaluate their ability to improve detection of frauds  on a dataset of financial transaction which is held by a national HSBC Payment Fraud Group. These large-scale real-world data consist of temporal patterns of customers’ digital journeys and payments. While the limited scale and granularity of public synthetic datasets restricts the generalisability of findings based on them to real-world settings, the data used in this study naturally incorporates the noise, outliers and irregularities that are common in production systems.

Our key findings are as follows.

\begin{itemize}
    \item Using time-encoder  with a lightweight LayerNorm-based integration of the time embeddings can improve the performance of a model for fraud-detection tasks.   
    \item 
    Event-level relative sinusoidal embeddings paired with a learned positional encoder deliver the highest performance on the test set, outperforming three classical baselines. This comparison varies the  timestamp scheme (absolute vs relative), the encoder (sinusoidal vs rotary),  and the presence or absence of a positional encoder.
\end{itemize}
Based on these findings we propose an algorithm which we call {\it FraudTransformer}, combining a vanilla GPT architecture with 
time and positional encoders.

This paper is organized as follows. Section~\ref{sec:relworks} surveys related work on transformer-based fraud detection. Section~\ref{sec:problem_stat} defines the problem. Section~\ref{sec:alg_mod} details our FraudTransformer  algorithm and GPT model. Section~\ref{sec:dataset} describes the dataset, and Section~\ref{sec:exp} presents the experiments and results. Section~\ref{sec:con} concludes and outlines directions for future work. Hyperparameter tuning results are deferred to the Appendix.


\section{Related Works} \label{sec:relworks}

Fraud detection methods typically fall into two main groups: classical supervised learning on tabular transaction data and deep learning for tabular inputs  For tabular supervised learning, gradient-boosted decision trees—most notably XGBoost and LightGBM—are strong, low-latency baselines because they exploit hand-engineered behavioral features; LightGBM remains a standard reference point for card and e-commerce fraud tasks \cite{ke2017lightgbm}. In tabular deep learning, multilayer perceptrons (MLPs) and autoencoders are common supervised and reconstruction-based options, while specialized models such as TabNet, which uses sequential feature attention to improve interpretability and often narrows the gap to boosted trees \cite{arik2021tabnet}, and TabTransformer, which applies self-attention to categorical embeddings \cite{huang2020tabtransformer}, have gained traction. In this work, we focus on transformers applications for fraud detection.

Transformer architectures have revolutionised the field of machine learning over the last 10 years.  While transformer models
are computationally expensive to train, once trained they can be deployed in real time. Fundamental to the advances of this approach is so-called attention, the ability of this architecture when considering a large sequence of inputs to decide what to pay attention to. 

Its strengths make transformers suitable for fraud detection (e.g. see ~\cite{zhang2023fata,padhi2021tabular,hu2023bert4eth,yang2023finchain}). Indeed,
 transformer architectures are increasingly central to fraud detection research, with investigations spanning encoder-only, decoder-only, and hierarchical designs. Encoder-type models, like BERT, are typically adapted through transfer learning; for instance, \cite{taneja2025fraud} showed that fine-tuning a pre-trained BERT on an online-recruitment corpus led to substantial gains in detection accuracy. Similarly, decoder-only transformers have been explored by tokenizing transactional events and pre-training GPT models on these sequences, leveraging their autoregressive nature to flag anomalous patterns, as demonstrated 
 in \cite{stein2024simple}. To capture richer structures in tabular data, hierarchical variants have been proposed, such as TabFormer, a BERT-based architecture that embeds individual transactions before aggregating them into table-level representations, achieving state-of-the-art performance \cite{padhi2021tabular}. Building on this, \cite{zhang2023fata} proposed FATA-trans to incorporate static fields—constant across all transactions—via separate encoders, further enhancing the ability of the model 
 to detect fraud. In the FATA-trans structure, inspired by sinusoidal positional embedding, a time-aware positional embedding based on absolute timestamps of transactions 
is added to embedding of transactions. 

 Motivated by the success of generative pretrained transformers (GPT) for large language models, GPT architectures have also been employed for social security fraud detection 
\cite{jin2025large}. 
A more detailed survey on transformer-based anomaly detection can be found
for example in \cite{ma2024research}.

To the best of our knowledge, using the relative time at event level in combination with a GPT model for financial fraud detection tasks has not yet been proposed. This paper fills that gap.

\section{Problem Statement}\label{sec:problem_stat}
Fraud detection requires capturing complex temporal and contextual dependencies across various data modalities, such as transactional metadata, user behaviour, and timestamps. Traditional machine learning models often struggle to process such intricate relationships.

While a GPT architecture is fast, by default all tokens are assigned the same importance. Although attention mechanisms are learnt to focus on the most important features, empirically, learning time and position-dependence in features is a challenge for GPT models~\cite{radford2018improving}. To address this issue, so-called positional encoders have been proposed. Positional encodings aid transformers in utilising positional information, originally regarding words in sequences.
There are two widely used positional encoding schemes for transformers, 
\begin{itemize}
    \item Sinusoidal positional encoding \cite{vaswani2017attention}; 
\item Rotary positional encoding \cite{su2024roformer}. 
\end{itemize}

A GPT transformer does not have an intrinsic notion of time and hence detecting time-dependent patterns tends to be difficult. To give the model temporal awareness, we use positional encoders but now as time encoders, replacing the usual position indices in the positional encoders with time identifiers (time-ids).  In practice, we pair the two encoding methods – sinusoidal and rotary – with one of the timestamps schemes below:
\begin{itemize}
    \item \textbf{Absolute scheme:} Each event has a unique timestamp in milliseconds, and we feed this absolute timestamp to the time encoders. 
\item \textbf{Relative scheme:} Instead of raw timestamps, we use the difference between timestamps of tokens or events as input to the time encoders.
\end{itemize}

The intuition behind using the Relative scheme is that of detecting patterns of time difference in the data. Also, the Absolute scheme expose calendar signals (hour-of-day, day-of-week, seasonality) and long-term trends, and let the model compare events across sequences on a shared external clock. Each account may have its own patterns, such as paying many bills at the beginning of the month and then reduced activity towards the end of the month. For detecting such patterns, rather than the Absolute timestamp, it is the time between transactions that should  contains (most of) the signal.

\begin{figure}
    \centering
    \includegraphics[width=0.95\linewidth]{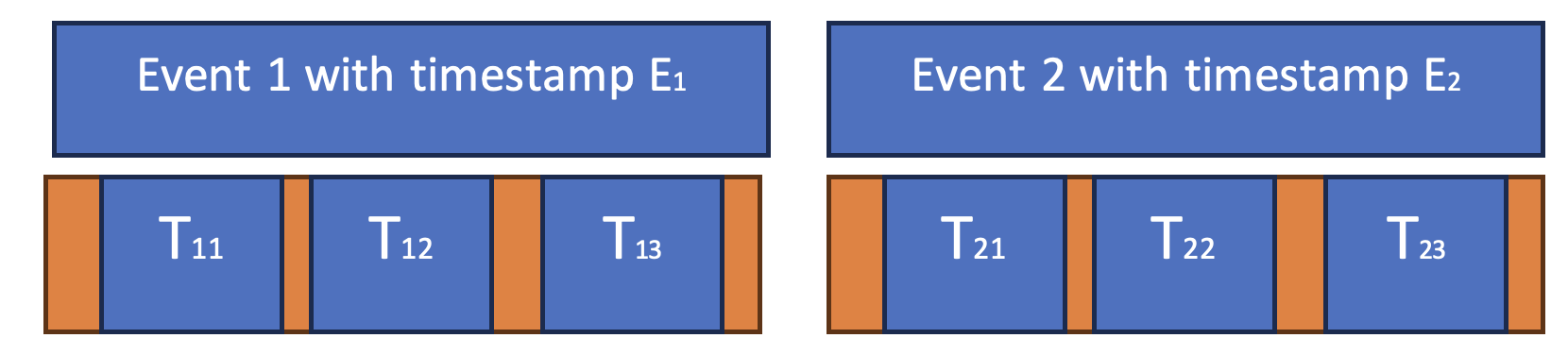}
    \caption{Event and Tokens Timestamps}
    \label{fig:event-label}
\end{figure}

Furthermore, regarding the Relative scheme, there are two options, a \textit{token-level} option and an \textit{event-level} option. To clarify these options, as in Figure \ref{fig:event-label}, suppose Event 1 has timestamp $E_1$ and tokens $T_{11}$,  $T_{12}$, and $T_{13}$,  while Event 2 has timestamp $E_{2}$ and tokens $T_{21}$, and $T_{23}$.  
In a token-level approach, we assign each token the timestamp from the corresponding event. Thus, each token in Event 1 is assigned timestamp $E_1$, and each token in Event 2 is assigned timestamp $E_2$. To compute the time difference for a token, we take the time difference to its previous token which is subsequently input to the time encoders. In the example above, the time difference for tokens $T_{12}$ and $T_{13}$ would be $0$, while the time difference for token $T_{21}$ is $E_2-E_1$; the time differences for $T_{22}$ and $T_{23}$ would again be $0$.
In the event-level approach, we use instead the time difference between the event to which the token belongs, and the previous event, as input to the time encoder. Now, $T_{21}$ would be assigned time difference $E_2-E_1$,  and $T_{22}$ as well as $T_{23}$ would also be assigned the time difference $E_2-E_1$.
 After a brief exploration, we decided to use the event-level approach.

\section{Dataset}\label{sec:dataset}
The HSBC Payment Fraud Group dataset used 
in this study consists of transactions and digital journeys with labelled fraud events. The evaluation dataset was sampled to conceal the fraud rate of the main data. While this implies 
that the performance of all models presented here may be different on datasets with other fraud rates, the primary aim of the study is to evaluate the efficacy of different time-embedding approaches for capturing nuanced patterns in temporally irregular events like payments. We employ this dataset to evaluate the pipelines discussed in the previous section, with the following specifications. 

\textbf{Features:} The HSBC Payment Fraud dataset contains more than 10 numerical features (e.g., transaction amount) and more than 10 categorical features (e.g., sort code); this comprehensive set of features based on events is in line of what is reported in the literature. Due to commercial sensitivity of this application domain we are unable to give a full account of all of the features. However, we highlight that in contrast to prior works that rely on the Tabformer dataset \cite{padhi2021tabular}, our approach leverages the complete set of information throughout the customer payment journey. To construct the consumers digital journey we consider around the last 500 events. 

\textbf{Data splits:} 
The dataset is constructed first using a temporal split to extract an independent test set, and a training set guaranteeing that 
the models are evaluated on future data.
Each of these sets are subsampled to allow scalability in training and inference and to conceal the fraud rate. The fraud
rate in the test data is around $43 \%$; this rate does not represent the fraud rate in the whole dataset but is instead obtained through
undersampling non-fraud cases.
An additional validation set for hyperparameter tuning is constructed by sampling from the training set.
This results in a total dataset  with $332{,}000$ samples, consisting of training $\approx$60 \% training, $\approx$ 30 \% validation, and 
$\approx$ 10 \% test samples.  

\section{Encoding process} 
We reduced the cardinality of categorical variables by grouping rare and low-risk events. Numerical variables were binned. The number of bins and bin edges were chosen to closely follow their underlying distribution while minimizing the number of bins.

Similar to \cite{stein2024simple}, we encode features for GPT model using a customized tokenization process with vocabulary size ($V$) around $4000$ tokens. We do not use sub-word tokenization; every category or bucket is treated as a single token. Then, we apply token embedding for tokens. We also tokenize each column separately, so identical strings in different columns become distinct tokens. For example, ``true'' in column A is a different token from ``true'' in column B. We also apply column embedding. In the following we provide more details regarding token and column embeddings.

In the following, we let $d_{\text{emb}}$ denote the embedding size of GPT model.

\paragraph{\textbf{Token embedding ($e^{\text{tok}}_t$)}.}
We use a learnable lookup table $W_{\text{tok}}\!\in\!\mathbb{R}^{V\times d_{\text{emb}}}$ to map token id $x_t\!\in\!\{0,\ldots,V{-}1\}$ to a vector:
\[
e^{\text{tok}}_t \;=\; W_{\text{tok}}[x_t].
\]
This contributes $V\!\cdot\! d_{\text{emb}}$ parameters and is trained end-to-end. We tie weights with the LM head (the output projection reuses $W_{\text{tok}}^\top$), so the encoder does not introduce an additional vocabulary-scale matrix. 

\paragraph{\textbf{Column embedding ($e^{\text{col}}_t$).}}
To disambiguate identical surface tokens that occur in different fields (e.g., \texttt{“100”} as an \emph{amount} vs.\ a \emph{merchant\_id}), we add a schema-aware embedding:
\[
e^{\text{col}}_t \;=\; W_{\text{col}}[c_t],\qquad
W_{\text{col}}\!\in\!\mathbb{R}^{C\times d_{\text{emb}}},
\]
where $c_t\!\in\!\{0,\ldots,C{-}1\}$ indexes the token’s source column and $C$ is the number of columns. This adds $C\!\cdot\! d_{\text{emb}}$ parameters and is learned jointly with the model. Intuitively, $e^{\text{col}}_t$  provides a field-specific signal that helps the attention layers learn different patterns for different columns of features, even when the raw tokens overlap.

The input embedding, which serves as the base representation for each token, is obtained by adding the token and column embeddings.

\begin{table*}[h]
  \centering
  \caption{Controlled ablation axes.  Only one factor is changed per row; all
           other settings match the full model.}
  \label{tab:ablation_axes}
  \begin{tabular}{@{}p{2.8cm}p{4.8cm}p{6.2cm}@{}}
    \toprule
    \textbf{Block tested} & \textbf{Main question to answer?} & \textbf{What is varied / removed} \\ \midrule
    Time embedding        & Is any explicit time signal needed? &
      Remove the entire time-encoder stack (“no-time” baseline) \\
    Positional encoding & Do sequence-order signals contribute beyond time stamps? &
       Remove the positional embeddings  
       \\
    Encoder type          & Extrapolation vs.\ locality trade-off &
      Swap the rotary and  
      the sinusoidal time encoders \\
    Timestamp scheme      &Absolute vs.\,Relative time scheme &
      Feed Absolute Unix time at event-level\\
      Normalization & Role of Layer-Norm & Remove the  LayerNorm block from time-encoders
    \\ \bottomrule
  \end{tabular}
\end{table*}

\section{Algorithm and Model}\label{sec:alg_mod}

\begin{figure}
    \centering
    \includegraphics[width=0.9\columnwidth]{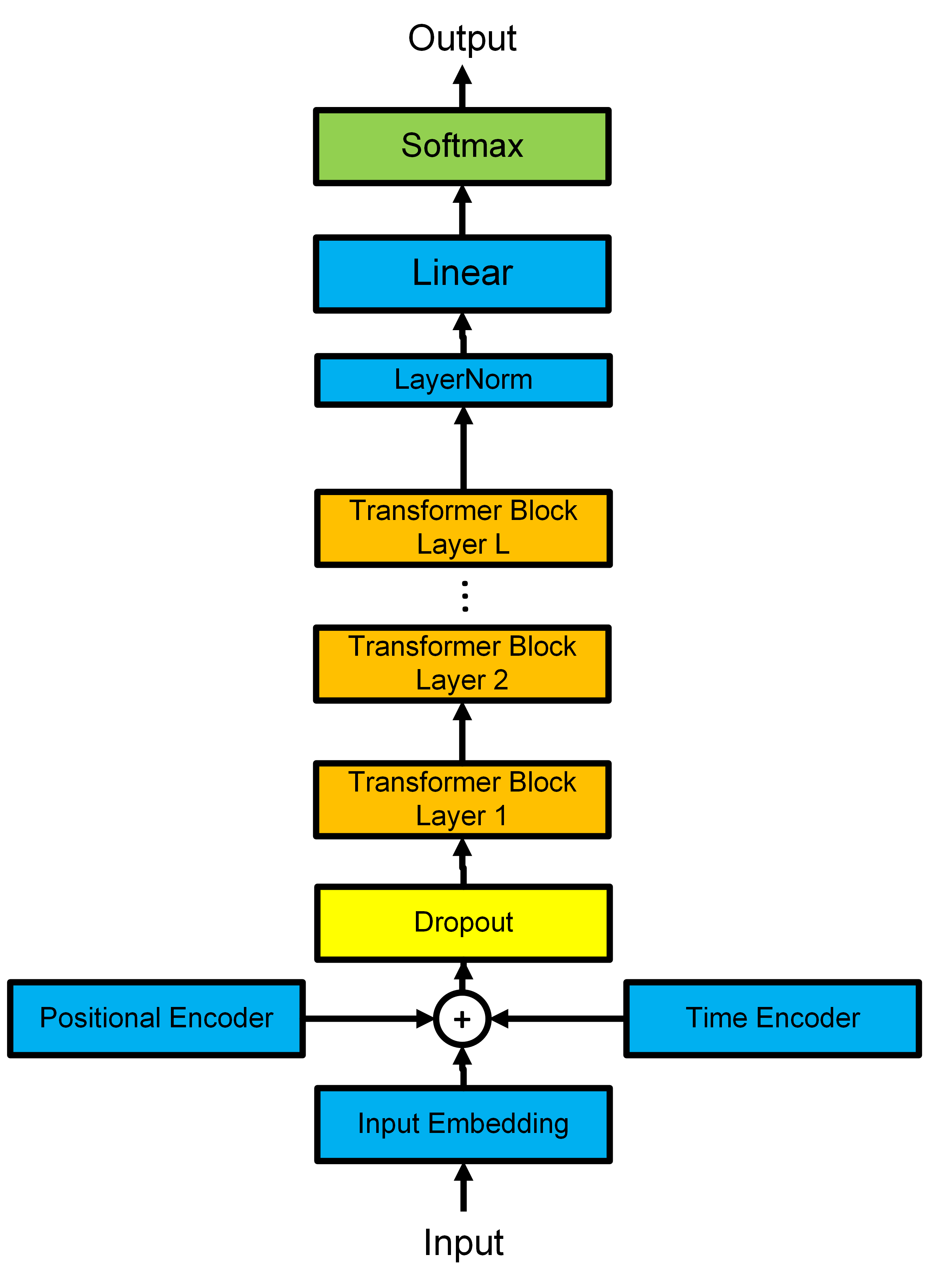}
    \caption{GPT Transformer with Time Encoder}
    \label{fig:Trnasformer}
\end{figure}

In this section, we introduce our model and algorithm for fraud detection using time-encoders and positional encoders based on absolute and relative timestamps. The diagram of our transformer block is shown in Figure \ref{fig:Trnasformer}. 

We let $L_{\max}$ denote the maximum sequence length (context length). We implement a 
vanilla GPT-style architecture \cite{radford2018improving} with 
$8$ heads,
$6$ layers, context length ($L_{\max}$) of $1024$, and embedding size 
($d_{\text{emb}}$)  of $512$, resulting in a model with around $20M$ parameters. A summary of the model configuration is provided in Table~\ref{tab:model_config}.

\begin{table}[htb]
  \centering
  \caption{Model configuration summary.}
  \label{tab:model_config}
  \begin{tabular}{@{}ll@{}}
    \toprule
    \textbf{Component} & \textbf{Setting} \\
    \midrule
    Architecture & Vanilla GPT-style~\cite{radford2018improving} \\
    Layers & $6$ Transformer blocks \\
    Attention heads & $8$ per layer \\
    Context length ($L_{\max}$) & $1024$ \\
    Embedding size ($d_{\text{emb}}$) & $512$ \\
    Parameter count & $\sim 20$M \\
    Vocabulary size & $\sim 4000$\\
    \bottomrule
  \end{tabular}
\end{table}

This model is augmented with  time and positional encoders. We train the model using the binary cross entropy loss function over the fraud and non-fraud classes.
The resulting algorithm, which we call \textit{FraudTransformer}, 
supports both \emph{Absolute} and \emph{Relative} timestamps; the attention mechanism in the GPT is unchanged.
The details are as follows.

\paragraph{Preliminaries.}
Let $t$  index tokens and $l$  index layers. Let $h_t^{(l)}\in\mathbb{R}^{d_{\text{emb}}}$ be the residual (pre-attention) hidden state in layer $l$ for token $t$.
Define a discrete time-id $\mathrm{id}_t\in \{0, 1, 2, \ldots\}  $ 
from wall-clock time or offsets:
\[
\mathrm{id}_t^{\text{abs}}= t_{\text{event}},\qquad
\mathrm{id}_t^{\text{rel}}= t_{\text{event}}-t_{\text{pre-event}},
\]
where $t_{\text{event}}{=t_{\text{event}}(t)}$ is the time of event to which the token $t$ belongs.
 Let $p_t\in\{0,1,\dots\}$ be the 
 position index of token $t$; often, this is different from $t_{\text{event}}(t)$. For vectors $a,b\in\mathbb{R}^n$, the Hadamard product is
$a\odot b = (a_1b_1,\dots,a_nb_n)$.

\begin{enumerate}
\item \textbf{Time encoding (additive features).}  For a given base frequency $f_{\rm base}$, which is a hyperparameter, we build 
$e_t\in\mathbb{R}^{d_{\text{emb}}}$ from $\mathrm{id}_t$ using one of the two options below.
\begin{itemize}
  \item \emph{Sinusoidal encoding (GPT-style Fourier features):} for even $d_{\text{emb}}$, let $\theta_i=f_{\text{base}}^{-2i/d_{\text{emb}}}$ with base frequency $f_{\text{base}}$ and $i=0,\dots,\tfrac{d_{\text{emb}}}{2}-1$, and set
  \[
  e_t^{\sin}=\big[\,\sin(\theta\odot \mathrm{id}_t)\ ;\ \cos(\theta\odot \mathrm{id}_t)\,\big]\in\mathbb{R}^{d_{\text{emb}}}.
  \] 
  \item \emph{Rotary-derived (used additively):}
  Assume $d_{\text{emb}}$ is even. Define the inverse-frequency values
  with base frequency $f_{\text{base}}$:
  \[
    \omega_i \;=\; f_{\text{base}}^{-\,\frac{2i}{d_{\text{emb}}}},\qquad i=0,\dots,\frac{d_{\text{emb}}}{2}-1.
  \]
  Form angles by taking the outer product with the time-ids:
  \[
    \phi_{t,i} \;=\; \mathrm{id}_t \,\omega_i,
  \]
  where $\mathrm{id}_t\in\{\mathrm{id}_t^{\text{abs}},\mathrm{id}_t^{\text{rel}}  \}.$ 
  For each 2D pair $(2i,2i\!+\!1)$, \emph{duplicate} the scalar time-id as a template $(x_{t,i},y_{t,i})=(\mathrm{id}_t,\mathrm{id}_t)$ and rotate it:
  \[
  \begin{bmatrix}
  e_{t,\,2i}\\ e_{t,\,2i+1}
  \end{bmatrix}
  =
  \begin{bmatrix}
  \cos\phi_{t,i} & -\sin\phi_{t,i}\\
  \sin\phi_{t,i} & \ \cos\phi_{t,i}
  \end{bmatrix}
  \begin{bmatrix}
  \mathrm{id}_t/f_{\text{base}}\\ \mathrm{id}_t/f_{\text{base}}
  \end{bmatrix}.
  \]
  Concatenating 
  these pairs over $i=0,\dots,\frac{d_{\text{emb}}}{2}-1$ in alternating order (even/odd slots) yields $e_t\in\mathbb{R}^{d_{\text{emb}}}$. 
\end{itemize}
Details on the hyperparameter tuning results are given in Section \ref{subsec:hyper}.

\item \textbf{Preparation (Normalization).} 
Apply LayerNorm \cite{ba2016layer} \emph{only} to the time embedding $e_t$ (with learnable parameters
$\beta_e,\tau_e$):
\[
\tilde e_t=\mathrm{LN}_{\beta_e,\tau_e}(e_t)=
\tau_e\odot\frac{e_t-\mu(e_t)}{\sqrt{\sigma^2(e_t)+\varepsilon}}+\beta_e,
\]

where  $\varepsilon>0$ is used for numerical stability and
\begin{equation}
    \begin{split}
         \mu(e_t)&=\frac{1}{d_{\text{emb}}}\sum_{i=1}^{d_{\text{emb}}} e_{t,i},\qquad \\
         \sigma^2(e_t)&=\frac{1}{d_{\text{emb}}}\sum_{i=1}^{d_{\text{emb}}}\bigl(e_{t,i}-\mu(e_t)\bigr)^2.  \end{split}
\end{equation}

\item \textbf{Learned Positional Encoders.} We use GPT-2–style learnable positional embeddings~\cite{hf-gpt2-docs-v4-38-0}. We maintain a parameter matrix
$\mathrm{EmbPos}\in\mathbb{R}^{L_{\max}\times d_{\text{emb}}}$ and perform a
row lookup by position index $p_t\in\{0,\dots,L_{\max}-1\}$:
\[
p_t^{\mathrm{learn}}=\mathrm{EmbPos}[p_t].
\]

We emphasize that no LayerNorm is applied to $p_t^{\mathrm{learn}}$. 

\item \textbf{Hidden-state update.}
Let $\lambda\in\mathbb{R}^{+}$ be a tunable parameter which represents time gain.
Update the residual stream and feed it to the unmodified block of transformer:
\begin{equation}\label{eq:tg}
    \bar h_t^{(1)} = h_t^{(1)} + p_t^{\mathrm{learn}} + \lambda\hat e_t,
\quad\text{with}\quad
h_t^{(1)} = e^{\text{tok}}_t + e^{\text{col}}_t,
\end{equation}
where $e^{\text{tok}}_t, e^{\text{col}}_t\in \mathbb{R}^{d_{\text{emb}}}$ are token and column embeddings, respectively and $p_t^{\mathrm{learn}}, \hat e_t \in \mathbb{R}^{d_{\text{emb}}}$. The Transformer blocks then proceed \emph{unchanged} with $\bar h_t^{(1)}$ as input embedding. 

\end{enumerate}


\begin{table}[htp]
  \centering
  \caption{Model variants considered in the time–encoder ablation study.
           Each variant toggles three binary choices:
           \emph{time scheme} (absolute vs.\ Relative),
           \emph{encoder type} (sinusoidal vs.\ rotary),
           and the presence of an additional \emph{positional} encoder.
           These
           {abbreviations} are used throughout the results section.}
  \label{tab:model_variants}
  \begin{tabular}{@{}llll@{}}
    \toprule
    \textbf{ID} & \textbf{Time encoder} & \textbf{Time scheme}  & \textbf{Positional encoder} \\ \midrule
    SA     & Sinusoidal &Absolute & \xmark \\
    SAP   & Sinusoidal &Absolute & \cmark \\
    SR     & Sinusoidal & Relative & \xmark \\
    SRP    & Sinusoidal & Relative & \cmark \\
    RA    & Rotary &Absolute     & \xmark \\
    RAP    & Rotary &Absolute    & \cmark \\
    RR    & Rotary     & Relative & \xmark \\
    RRP    & Rotary    & Relative  & \cmark \\ \bottomrule
  \end{tabular}
\end{table}

\begin{table*}[htp]
  \centering
  \caption{Baseline models and Transformer variants evaluated on the HSBC Payment Fraud Group’s test set (mean\,$\pm$\,standard deviation over three runs; higher is better{\color{black}; denoted by $\uparrow$}). \textbf{TE}=time encoder, \textbf{PE}=positional encoder. Time–encoder codes follow Table\,\ref{tab:model_variants}: \textbf{S}=sinusoidal, \textbf{R}=rotary; \textbf{A}=absolute timestamps, \textbf{R}=relative. {\color{black} The $\Delta$PRAUC and $\Delta$AUROC columns report the difference of the PRAUC and AUROC, respectively, to the  mean of the Vanilla baseline (no TE, no PE).}}
  \label{tab:baseline_results}

  \begin{tabular}{@{}ccccc@{}}
    \toprule
    \textbf{Model} & \textbf{PRAUC $\uparrow$} & \textbf{AUROC $\uparrow$} & \makecell{\textbf{$\Delta$PRAUC}} & \makecell{\textbf{$\Delta$AUROC}} \\
    \midrule
    \multicolumn{5}{l}{\textit{Feature-based baselines}} \\
    \quad Logistic Regression                & $0.92491\pm 0.00000$ & $0.93784\pm0.00000$ & $+0.00453$ & $+0.00182$ \\
    \quad XGBoost                            & $0.93447\pm 0.00000$ & $0.95894\pm 0.00000$ & $+0.01409$ & $+0.02292$ \\
    \quad LightGBM                           & $0.94536\pm 0.00000$ & $0.96212\pm 0.00000$ & $+0.02498$ & $+0.02610$ \\
    \addlinespace
    \multicolumn{5}{l}{\textit{Transformer – no positional encoder}} \\
    \quad Vanilla (baseline) \;(no TE, no PE)               & $0.92038\pm 0.00019$ & $0.93602\pm 0.00067$ & $+0.00000$ & $+0.00000$ \\
    \quad \textit{RA}  (rotary, Abs)        & $0.91278\pm0.00010$ & $0.93003\pm 0.00021$ & $-0.00760$ & $-0.00599$ \\
    \quad \textit{RR}  (rotary, Rel)        & $0.92145\pm 0.00049$ & $0.93989\pm 0.00072$ & $+0.00107$ & $+0.00387$ \\
    \quad \textit{SA}  (sinusoidal, Abs)    & $0.91539\pm0.00043$ & $0.93034\pm 0.00079$ & $-0.00499$ & $-0.00568$ \\
    \quad \textit{SR}  (sinusoidal, Rel)    & $0.92477\pm 0.00012$ & $0.94212 \pm 0.00067$ & $+0.00439$ & $+0.00610$ \\
    \addlinespace
    \multicolumn{5}{l}{\textit{Transformer – with positional encoder}}  \\
    \quad \textit{NTE} (no TE, with PE)     & $0.94689\pm 0.00109$ & $0.95664\pm 0.00064$ & $+0.02651$ & $+0.02062$ \\
    \quad \textit{RAP} (rotary, Abs)        & $0.94436\pm 0.00080$ & $0.95539\pm 0.00022$ & $+0.02398$ & $+0.01937$ \\
    \quad \textit{RRP} (rotary, Rel)        & $0.95610\pm 0.00118$ & $0.96497\pm 0.00041$ & $+0.03572$ & $+0.02895$ \\
    \quad \textit{SAP} (sinusoidal, Abs)    & $0.94713\pm 0.00042$ & $0.95639\pm 0.00093$ & $+0.02675$ & $+0.02037$ \\
    \quad \textbf{SRP} (sinusoidal, Rel)    & $\pmb{0.95816\pm 0.00092}$ & $\pmb{0.96723\pm0.00011}$ & $\pmb{+0.03778}$ & $\pmb{+0.03121}$ \\
    \addlinespace
    \multicolumn{5}{l}{\textit{SRP – without LayerNorm (WOL)}}\\
    \quad SRP/WOL                            & $0.95304\pm 0.00102$ & $0.96301\pm0.00091$ & $+0.03266$ & $+0.02699$ \\
    \bottomrule
  \end{tabular}
\end{table*}

\section{Experiments}\label{sec:exp}

\subsection{Evaluation Metrics}
We report \textbf{PRAUC} (area under the precision--recall curve) and \textbf{AUROC} (area under the receiver--operating characteristic curve). PRAUC is particularly informative for fraud detection because  the class imbalance ratio can be very high. More details can be found for example in \cite{davis2006relationship}.

\subsection{Baselines} \label{sec:baselines}
 Here we detail the baselines against which we compare our new architecture.

\paragraph{LightGBM} A tuned gradient--boosted decision tree ensemble, \cite{ke2017lightgbm}, trained on the same partition of the training dataset. For a fair comparison, we use the same set of features in LightGBM  as in the transformer models.

\paragraph{GPT (``no time'').} A transformer encoder identical to our proposed architecture except that all Absolute and Relative time encodings and positional encoding are removed. We include two versions, with and without positional encoding. These two baselines isolate the impact of explicit time modelling.

\paragraph{XGBoost.} 
We also train an \emph{eXtreme Gradient Boosting} ensemble \cite{chen2016xgboost} on the same training split.  
For direct comparability, the feature set fed to XGBoost is identical to that of LightGBM.

\paragraph{Logistic~Regression.} 
As a strong linear baseline, we fit an $\ell_{2}$-regularized logistic regression model \cite{cox1958regression} on the same feature matrix.  
The inverse regularization strength $C$ and class‐weight ratio are selected by grid search on the validation fold, while all features are standardised to zero mean and unit variance.  
Apart from these tuned hyperparameters, we keep the scikit-learn \cite{sklearnlogreg} defaults, ensuring that the only difference from the tree-based models is the underlying learning algorithm.

\subsection{Hyperparameter Tuning} \label{subsec:hyper}
We consider the time gain $\lambda$, as defined in \eqref{eq:tg}, and the base frequency, $f_{\text{base}}$, of the sinusoidal and rotary positional encoders as tunable hyperparameters.  For a fair comparison, for every hyperparameter pair, the model is trained for a single epoch and assessed on the validation set at the end of first epoch. We consider $\lambda=0.01$ and $f_{\text{base}}=10^7$ for our main experiments. The results are found in the Appendix~\ref{sec:hyper_tuning}.

\subsection{Fraud Types}

We consider multiple fraud subtypes (such as scams \cite{hsbcCommonScams2025}, and account takeover \cite{hsbc_account_takeover_2025}). Although our model is trained as a binary classifier (fraud vs. non-fraud), we examine whether its performance varies by subtype. For this purpose, we create subtype-specific test sets that include only non-fraud transactions and one fraud subtype, excluding all other fraud labels. We then evaluate the fixed model (our best model) on each subset and report AUROC and PR-AUC. For example, for the scam subset, the test set contains only scam and non-fraud cases; the resulting AUROC and PR-AUC quantify how well the model separates scams from non-fraud transactions.
\subsection{Ablation Study} \label{subsec:ablation}

Our ablation study is designed to isolate the contribution of each component in the temporal encoding stack. We consider three factors:
(a) the \emph{time encoder family} (sinusoidal vs.\ rotary),
(b) the \emph{time scheme} (absolute vs.\ event-level relative), (c) the presence of an additional \emph{positional} encoder (learned vs.\ none) and (d) the presence of Layer-Norm layer (Normalize vs.\ none) detailed in Table~\ref{tab:ablation_axes}.


We keep the architecture, optimization settings, data splits, and preprocessing fixed across variants. We evaluate each model on the held-out test set. We report PRAUC and AUROC. The combinations of time and positional encoders  which we investigate are shown in Table~\ref{tab:model_variants}. The results for this study are provided in next section.

\subsection{Results}

\begin{table*}
\centering
\caption{Performance comparison of Vanilla vs. SRP across fraud types. Improvements are absolute gains (SRP - Vanilla).}
\label{tab:subtype}
\begin{tabular}{lcccccc}
\toprule
\multirow{2}{*}{\textbf{Fraud Type}} & 
\multicolumn{2}{c}{\textbf{Vanilla GPT}} & 
\multicolumn{2}{c}{\textbf{SRP}} & 
\multicolumn{2}{c}{\textbf{Improvement}} \\
\cmidrule(lr){2-3} \cmidrule(lr){4-5} \cmidrule(lr){6-7}
 & PRAUC & AUROC & PRAUC & AUROC & $\Delta$ PRAUC & $\Delta$ AUROC \\
\midrule
Scam        & 0.8806 & 0.9406 & 0.9337 & 0.9704 & +0.0531 & +0.0297 \\
Other Frauds & 0.8477 & 0.9374 & 0.8958 & 0.9627 & +0.0481 & +0.0254 \\
ATO         & 0.8198 & 0.9329 & 0.8737 & 0.9591 & +0.0538 & +0.0263 \\
\bottomrule
\end{tabular}
\end{table*}

Table~\ref{tab:baseline_results} contrasts the proposed \emph{GPT with Time encoder and Positional encoder} model with different positional encoders with the three classical feature-based learners  from Section \ref{sec:baselines} and with different reduced GPT variants.
 All runs used fixed train/validation/test splits and deterministic hyperparameters \footnote{We select no row/feature subsampling: LightGBM bagging\_fraction=1.0, feature\_fraction=1.0; XGBoost subsample=1.0, colsample\_bytree=1.0.}. Consequently, training and evaluation are deterministic, yielding identical results across repeated runs. Features are from a single time snapshot for these comparisons. Accordingly, the standard deviation of the results for the three baselines are zero. The results are shown in Table \ref{tab:baseline_results}.
 
 Among the non-transformer baselines, LightGBM achieves the strongest performance but is outperformed by the transformer-based models with positional encoding (e.g., GPT-style) and time encoding (relative scheme) on both evaluation metrics.

To assess the contribution of LayerNorm (Step 2 of the FraudTransformer algorithm; Section~\ref{sec:alg_mod}), we perform an ablation in which SRP omits LayerNorm on the time-encoder output.
The result (SRP/WOL)  in Table~\ref{tab:baseline_results}
 shows a slight but significant deterioration.

Regarding the fraud-types, we evaluated the performance of our best model  (SRP) against a vanilla GPT for scams, account takeover (ATO), and a category of other types of frauds. The results are reported in Table~\ref{tab:subtype}; the SRP model improves on the vanilla model for all three fraud types considered.

\subsection{Discussion}

Our strongest configuration is sinusoidal time embeddings with 
learned positional embeddings, event-level Relative, and LayerNorm, appearing as SRP in Table \ref{tab:baseline_results}. 
 In the following, we provide a more detailed discussion based on our observations in experiments. 

 {\color{black}\textbf{PRAUC Values:} The PRAUC baseline is the proportion of positive (that is here, fraudulent) entries. Hence with a $43\%$ fraud rate in our evaluation set, the random baseline is $ 0.43$, which may make the  PRAUC values appear to be high. Therefore, instead of reporting absolute PRAUC, we emphasise  relative differences between models on the same test set. For this purpose the $\Delta$ of PRAUC with respect to baseline, Vanilla GPT, is reported in Table~\ref{tab:baseline_results}.}

\textbf{Time and positional encoding (SRP) vs Vanilla GPT model:} The model with a relative time embedding strongly outperforms the baseline on both PRAUC and AUROC. This strongly demonstrates that incorporating time embedding features and positional features into the model improves the performance of the model. 
 
\textbf{Sinusoidal vs Rotary Time embedding:} We observe relatively similar performance between these two schemes when holding all other variables constant. Sinusoidal encoding has a modest advantage over Rotary encoding. This pattern is robust to both choice of performance measure (AUROC vs PRAUC), and relative or absolute time embeddings in this dataset. 

\textbf{Relative vs Absolute Time encoding:} As observed in Table~\ref{tab:baseline_results}, the relative time encoding options outperform the absolute time encoding options in both AUROC and PRAUC. 

\textbf{Positional Encoding:} Adding only a positional encoder yields a modest but consistent lift over the feature-based baselines, confirming that sequential structure carries information not captured by raw features alone.

\textbf{LayerNorm:} Table~\ref{tab:baseline_results} demonstrates that adding LayerNorm to SRP 
improves performance and lowers the standard deviation relative to SRP without LayerNorm.

\textbf{Fraud Sub-types:} As shown in Table~\ref{tab:subtype}, SRP improves detection across all fraud subtypes, with the largest gains in PRAUC on account takeover (ATO) fraud and in AUROC on scam fraud, relative to the vanilla transformer. Because ATO frauds happen over short temporal windows, the relative-time modeling in the SRP model appears captures these transient patterns more effectively than the vanilla model. This improvement is particularly encouraging as ATO is the type of fraud which is most difficult for the Vanilla model to detect in this data set.

\section{Conclusion and Future Works}\label{sec:con}

We have introduced \textit{FraudTransformer}, a GPT-style sequence model that is made explicitly time-aware through sinusoidal or rotary \emph{time encoders} and an additional learned positional encoder.  On a large real-world banking dataset held by a national HSBC Payment Fraud Group, the full variant that combines relative, event-level sinusoidal time embeddings with positional encoder consistently outperforms three strong feature-based baselines: Logistic Regression, XGBoost and LightGBM.  The gains hold across both AUROC and PRAUC, confirming that fine-grained temporal structure carries signal beyond what raw features capture.

In practice, FraudTransformer remains computationally lightweight, requiring only a single training epoch for hyperparameter sweeps. {\color{black} Moreover, FraudTransformer} 
delivers stable improvements that translate into fewer missed frauds and lower false-alarm rates.  

{\color{black} Some improvements are possible. This work trains the model end-to-end with a binary cross-entropy objective for sequence-level fraud classification. In future work, we{\color{black} suggest} to pre-train the GPT backbone using a next-token prediction objective~\cite{wang2024emu3}, with the expectation that the token positions will be learnt more accurately. While our current task is binary fraud vs. non-fraud, we could extend the model to multi-class prediction over fraud subtypes in future works. Finally, our approach can be extended to new GPT architectures which support larger context lengths.}


\section*{DISCLAIMER}
This paper was prepared for information purposes, and is not a product of HSBC or its affiliates. Neither HSBC nor any of its affiliates make any explicit or implied representation or warranty and none of them accept any liability in connection with this paper but not limited to, the completeness, accuracy, reliability of information contained herein and the potential legal, compliance, tax or accounting effects thereof. Copyright HSBC Group {\color{black} 2025}. The content of this manuscript was reviewed by HSBC’s publication clearance team to ensure no disclosure of sensitive data. 
\bibliographystyle{plain}
\bibliography{sample_ref}

\begin{thebibliography}{10}

\bibitem{hf-gpt2-docs-v4-38-0}
Open{AI} {GPT2,T}ransformers v4.38.0 documentation.
\newblock \url{https://huggingface.co/docs/transformers/v4.38.0/en/model_doc/gpt2}.

\bibitem{hsbc_account_takeover_2025}
Account {T}akeover – {HSBC UK B}usiness {B}anking.
\newblock \url{https://www.business.hsbc.uk/en-gb/account-takeover}, August 2025.
\newblock Accessed: 2025-08-31.

\bibitem{hsbcCommonScams2025}
Common scams.
\newblock \url{https://www.hsbc.co.uk/help/security-centre/fraud-guide/common-scams/}, 2025.
\newblock Accessed: 2025-08-31.

\bibitem{arik2021tabnet}
Sercan~{\"O} Arik and Tomas Pfister.
\newblock Tabnet: Attentive interpretable tabular learning.
\newblock In {\em Proceedings of the AAAI Conference on Artificial Intelligence}, volume~35, pages 6679--6687, 2021.

\bibitem{ba2016layer}
Jimmy~Lei Ba, Jamie~Ryan Kiros, and Geoffrey~E Hinton.
\newblock Layer normalization.
\newblock {\em arXiv preprint arXiv:1607.06450}, 2016.

\bibitem{chen2016xgboost}
Tianqi Chen and Carlos Guestrin.
\newblock Xgboost: A scalable tree boosting system.
\newblock In {\em Proceedings of the 22nd {ACMSIGKDD} International Conference on Knowledge Discovery and Data Mining}, pages 785--794, 2016.

\bibitem{cox1958regression}
David~R Cox.
\newblock The regression analysis of binary sequences.
\newblock {\em Journal of the Royal Statistical Society Series B: Statistical Methodology}, 20(2):215--232, 1958.

\bibitem{davis2006relationship}
Jesse Davis and Mark Goadrich.
\newblock The relationship between precision-recall and {ROC} curves.
\newblock In {\em Proceedings of the 23rd International Conference on Machine learning}, pages 233--240, 2006.

\bibitem{hilal2022financial}
Waleed Hilal, S~Andrew Gadsden, and John Yawney.
\newblock Financial fraud: a review of anomaly detection techniques and recent advances.
\newblock {\em Expert Systems With Applications}, 193:116429, 2022.

\bibitem{hu2023bert4eth}
Sihao Hu, Zhen Zhang, Bingqiao Luo, Shengliang Lu, Bingsheng He, and Ling Liu.
\newblock Bert4eth: {A} pre-trained transformer for ethereum fraud detection.
\newblock In {\em Proceedings of the ACM Web Conference 2023}, pages 2189--2197, 2023.

\bibitem{huang2020tabtransformer}
Xin Huang, Ashish Khetan, Milan Cvitkovic, and Zohar Karnin.
\newblock Tabtransformer: Tabular data modeling using contextual embeddings.
\newblock {\em arXiv preprint arXiv:2012.06678}, 2020.

\bibitem{jin2025large}
Ruohua Jin, Yanyan Wen, Tao Liu, and Cong Shen.
\newblock Large language models for fraud detection.
\newblock In {\em 2025 8th International Conference on Artificial Intelligence and Big Data (ICAIBD)}, pages 441--446. IEEE, 2025.

\bibitem{ke2017lightgbm}
Guolin Ke, Qi~Meng, Thomas Finley, Taifeng Wang, Wei Chen, Weidong Ma, Qiwei Ye, and Tie-Yan Liu.
\newblock Lightgbm: A highly efficient gradient boosting decision tree.
\newblock {\em Advances in Neural Information Processing Systems}, 30, 2017.

\bibitem{ma2024research}
Mingrui Ma, Lansheng Han, and Chunjie Zhou.
\newblock Research and application of {T}ransformer based anomaly detection model: A literature review.
\newblock {\em arXiv preprint arXiv:2402.08975}, 2024.

\bibitem{padhi2021tabular}
Inkit Padhi, Yair Schiff, Igor Melnyk, Mattia Rigotti, Youssef Mroueh, Pierre Dognin, Jerret Ross, Ravi Nair, and Erik Altman.
\newblock Tabular transformers for modeling multivariate time series.
\newblock In {\em ICASSP 2021-2021 IEEE International Conference on Acoustics, Speech and Signal Processing (ICASSP)}, pages 3565--3569. IEEE, 2021.

\bibitem{radford2018improving}
Alec Radford, Karthik Narasimhan, Tim Salimans, Ilya Sutskever, et~al.
\newblock Improving language understanding by generative pre-training.
\newblock 2018.

\bibitem{sklearnlogreg}
{scikit-learn developers}.
\newblock Logisticregression — scikit-learn documentation.
\newblock \url{https://scikit-learn.org/stable/modules/generated/sklearn.linear_model.LogisticRegression.html}.
\newblock Accessed: 2025-08-27.

\bibitem{stein2024simple}
Alex Stein, Samuel Sharpe, Doron Bergman, Senthil Kumar, C~Bayan Bruss, John Dickerson, Tom Goldstein, and Micah Goldblum.
\newblock A simple baseline for predicting events with auto-regressive tabular transformers.
\newblock {\em arXiv preprint arXiv:2410.10648}, 2024.

\bibitem{su2024roformer}
Jianlin Su, Murtadha Ahmed, Yu~Lu, Shengfeng Pan, Wen Bo, and Yunfeng Liu.
\newblock Roformer: {E}nhanced transformer with rotary position embedding.
\newblock {\em Neurocomputing}, 568:127063, 2024.

\bibitem{taneja2025fraud}
Khushboo Taneja, Jyoti Vashishtha, and Saroj Ratnoo.
\newblock Fraud-{BERT}: transformer based context aware online recruitment fraud detection.
\newblock {\em Discover Computing}, 28(1):1--16, 2025.

\bibitem{vaswani2017attention}
Ashish Vaswani, Noam Shazeer, Niki Parmar, Jakob Uszkoreit, Llion Jones, Aidan~N Gomez, {\L}ukasz Kaiser, and Illia Polosukhin.
\newblock Attention is all you need.
\newblock {\em Advances in Neural Information Processing Systems}, 30, 2017.

\bibitem{wang2024emu3}
Xinlong Wang, Xiaosong Zhang, Zhengxiong Luo, Quan Sun, Yufeng Cui, Jinsheng Wang, Fan Zhang, Yueze Wang, Zhen Li, Qiying Yu, et~al.
\newblock Emu3: Next-token prediction is all you need.
\newblock {\em arXiv preprint arXiv:2409.18869}, 2024.

\bibitem{yang2023finchain}
Xinze Yang, Chunkai Zhang, Yizhi Sun, Kairui Pang, Luru Jing, Shiyun Wa, and Chunli Lv.
\newblock Finchain-{BERT}: {A} high-accuracy automatic fraud detection model based on nlp methods for financial scenarios.
\newblock {\em Information}, 14(9):499, 2023.

\bibitem{zeng2023transformers}
Ailing Zeng, Muxi Chen, Lei Zhang, and Qiang Xu.
\newblock Are transformers effective for time series forecasting?
\newblock In {\em Proceedings of the Thirty-Seventh AAAI Conference on Artificial Intelligence and Thirty-Fifth Conference on Innovative Applications of Artificial Intelligence and Thirteenth Symposium on Educational Advances in Artificial Intelligence}, pages 11121--11128, 2023.

\bibitem{zhang2023fata}
Dongyu Zhang, Liang Wang, Xin Dai, Shubham Jain, Junpeng Wang, Yujie Fan, Chin-Chia~Michael Yeh, Yan Zheng, Zhongfang Zhuang, and Wei Zhang.
\newblock Fata-trans: Field and time-aware transformer for sequential tabular data.
\newblock In {\em Proceedings of the 32nd ACM International Conference on Information and Knowledge Management}, pages 3247--3256, 2023.

\end{thebibliography}
\newpage

\clearpage

\appendix







\section*{Appendix: Hyperparameter Tuning Results}\label{sec:hyper_tuning}
Table \ref{tab:hypersinusoidal} and Table \ref{tab:hyperrotary} show the hyperparameter tuning results for the RAP, RRP, SAP and SRP schemes. The  boldface parameters are selected as hyperparameters and are then also applied in the RA, RR, SA and SR schemes.

\medskip 
\begin{table}[b]
    \centering
        \caption{Hyperparameter Tuning for the Rotary Scheme, for Relative and Absolute time encoding. The best performing choice is in bold.}
    \label{tab:hypersinusoidal}
   \begin{tabular}{l r r r}
\hline
Encoding & $f_{\text{base}}$ & $\lambda$ & PRAUC (at epoch 1) \\
\hline
\multicolumn{4}{l}{\textbf{Rotary (Relative)}} \\
RRP & 100000000 & 0.01 & 0.95602  \\
RRP & 10000000 & 0.01 & \textbf{0.95615}  \\
RRP & 1000000 & 0.01 & 0.95498  \\
RRP & 100000 & 0.01 & 0.95209  \\
RRP & 1000 & 0.01 & 0.95070  \\
RRP & 10 & 0.01 & 0.94543 \\
RRP & 1 & 0.01 & 0.94427 \\
RRP & 100000000 & 0.1 & 0.94004  \\
RRP & 10000000 & 0.1 & 0.94304  \\
RRP & 1000000 & 0.1 & 0.94277  \\
RRP&  10000 & 0.1 & 0.93950  \\
RRP & 1000 & 0.1 & 0.92553  \\
RRP & 10 & 0.1 & 0.92564 \\
RRP & 1 & 0.1 & 0.91268 \\
\hline
\multicolumn{4}{l}{\textbf{Rotary (Absolute)}} \\
RAP & 100000000 & 0.01 & 0.94425  \\
RAP & 10000000 & 0.01 & 0.94430  \\
RAP & 1000000 & 0.01 & 0.94235  \\
RAP & 100000 & 0.01 & 0.94201  \\
RAP & 1000 & 0.01 & 0.93931  \\
RAP & 10 & 0.01 & 0.93360  \\
RAP & 1 & 0.01 & 0.92663  \\
RAP & 100000000 & 0.1 & 0.93936  \\
RAP & 10000000 & 0.1 & 0.93931  \\
RAP & 1000000 & 0.1 & 0.93054 \\
RAP & 100000 & 0.1 & 0.93226  \\
RAP & 1000 & 0.1 & 0.92893  \\
RAP & 10 & 0.1 & 0.92001  \\
RAP & 1 & 0.1 & 0.91676  \\
\hline
\end{tabular}
\end{table}

\medskip 
\begin{table}[b]
    \centering
        \caption{Hyperparameter Tuning for the Sinusoidal Scheme, for Relative and Absolute time encoding. The best performing choice is in bold.}
    \label{tab:hyperrotary}
   \begin{tabular}{l r r r}
\hline
Encoding & $f_{\text{base}}$ & $\lambda$ & PRAUC (at epoch 1) \\
\hline
\multicolumn{4}{l}{\textbf{Sinusoidal (Relative)}} \\
SRP & 100000000 & 0.01 & 0.95705 \\
SRP & 10000000  & 0.01 & \textbf{0.95820} \\
SRP & 1000000   & 0.01 & 0.95597 \\
SRP & 100000    & 0.01 & 0.95399 \\
SRP & 1000      & 0.01 & 0.95144 \\
SRP & 10        & 0.01 & 0.94041 \\
SRP & 1         & 0.01 & 0.93827 \\
SRP & 100000000 & 0.1  & 0.94114 \\
SRP & 10000000  & 0.1  & 0.94403 \\
SRP & 1000000   & 0.1  & 0.94336 \\
SRP & 10000     & 0.1  & 0.94050 \\
SRP & 1000      & 0.1  & 0.92633 \\
SRP & 10        & 0.1  & 0.92124 \\
SRP & 1         & 0.1  & 0.91368 \\
\hline
\multicolumn{4}{l}{\textbf{Sinusoidal (Absolute)}} \\
SAP & 100000000 & 0.01 & 0.94525  \\
SAP & 10000000  & 0.01 & 0.94730  \\
SAP & 1000000   & 0.01 & 0.94335  \\
SAP & 100000    & 0.01 & 0.94371  \\
SAP & 1000      & 0.01 & 0.94330  \\
SAP & 10        & 0.01 & 0.93469  \\
SAP & 1         & 0.01 & 0.92797  \\
SAP & 100000000 & 0.1  & 0.94036  \\
SAP & 10000000  & 0.1  & 0.94131  \\
SAP & 1000000   & 0.1  & 0.93154  \\
SAP & 100000    & 0.1  & 0.93326  \\
SAP & 1000      & 0.1  & 0.93293  \\
SAP & 10        & 0.1  & 0.92111  \\
SAP & 1         & 0.1  & 0.91776  \\
\hline
\end{tabular}

\end{table}

\end{document}